# Finetuning an LLM on Contextual Knowledge of Classics for Q&A

Shane Storm Strachan

*ssstra01@louisville.edu*

**Abstract**

The open-source publishing of large language models (LLMs) has created many possibilities for how anyone who understands language and has access to a computer can interact with significant tools of artificial intelligence, particularly in the context of learning and knowledge dissemination. However, the utility of these models in specialized fields like Classics is still largely unexplored. This project is an attempt to merge the knowledge of Classics with the capabilities of artificial intelligence by finetuning an LLM to cater to the specific needs of learners and professionals. The goal of this project is to develop an LLM that not only reproduces contextual knowledge accurately but also exhibits a consistent "personality"–and, indeed, has consistent propriety–to appeal to a diverse audience who possess differing levels of knowledge. A significant portion of this project was dedicated to refining the dataset, following the principle of "garbage in, garbage out," to ensure the model generates relevant, useful, and creative responses when given a prompt (a statement, question, or single word). After training and evaluation, my model's ability to handle a vast array of different types of inputs and prompting exceeded my original expectations for a 355M parameter model, though its occasional hallucinations (especially when set with a high temperature), particularly in its assertions about historical events or its own identity, make it seem somewhat capricious and more work in the form of continuous finetuning will be undertaken.

***Index Terms:*** *large language model, finetuning, GPT2, transformers and alethiology, natural language processing, question and answering, Classics and AI*

1. Introduction

The publishing of open source Large Language Models (LLMs) built on transformer technology, exemplified by the GPT series as well as LLaMa, has marked a significant milestone in the field of artificial intelligence, especially with regard to its potential for knowledge dissemination and better access to materials and guided teaching, and they often can perform particularly well on elementary knowledge (rather than more advanced, critical approaches to knowledge, or obscure or "controversial" knowledge). However, the application of these models in specialized academic disciplines is relatively uncharted, especially regarding Classics (indeed, partially due to the inherent need for it to have an understanding of multiple languages including Latin, process hundreds of years' worth of data, and the need to be able to historically contextualize). This project is an attempt to finetune an LLM specifically for the Classics discipline, aiming to make it a valuable resource for a diverse audience, ranging from beginners to scholars in the field.

The initial concept of this project revolved around introducing and classifying hierarchical knowledge in an LLM based on concepts of alethiology. However, it soon became apparent that this approach was impractical with the current constraints of transformer architecture, as well as my limitations in computational resources, financial investment, and time. Additionally, I abandoned the idea of employing a quantized model for several reasons, most importantly due to their inherent faults involved with finetuning (although recent advancements in the field suggest this is changing rapidly).

This project was undertaken using consumer-grade hardware, specifically an Nvidia 2080 Ti, because of my own limitations as mentioned, but also to demonstrate the feasibility and potential of conducting advanced research with accessible equipment and not needing to use cloud-based GPUs and their related issues of security and cost. This project, therefore, is at the intersection of technology, classical education, and questions regarding knowledge and truth and the ethics of LLMs and their myriad uses. This project is an endeavor to create a tool that is not only academically rigorous but also engaging and accessible, attempting to ensure that the model's responses are "critical" (logical, and philosophically aware of politics and power dynamics) and appropriately contextualized within the framework of Classics.

2. **Related Work**

LLMs operate within a dynamic and challenging context, particularly concerning their nature, their potential, and their popular use in industry, as exemplified by the multibillion dollar companies involved in their creation. As a result, these models and their reception by society has generated profound dialogue about the nature of knowledge construction and the reliability of generative AI and how it will be used by agents throughout the world with a myriad of different agendas. Indeed, the focus on the transformer architecture has revealed significant challenges that remain in generating and maintaining control and trust of information. One of the most prominent issues in this field is the phenomenon of hallucinations, or the generation of syntactically correct but factually incorrect or seemingly pathological text. However, the challenges extend beyond this, encompassing issues like outdated information, biases, inconsistent responses, and the potential to upend many aspects of society, from education, to business, to politics and the notion of human rights.

Recent scholarship has aimed at enhancing the capability of transformers and LLMs. A common theme among these studies is the implementation of fact-checking mechanisms or context "chains" to improve accuracy and do a sort of self-moderating process. For example, Li (2023) shows that targeted data can rival larger models in effectiveness. Indeed, Stanford's release of Alpaca (2023) suggests that impressive finetuning can be achieved with less computational power or data. Dhuliawala (2023) introduces the "Chain-of-Verification (CoVe)" concept, attempting to counter hallucinations and "erroneous" responses. Furthermore, Wan (2023) proposes domain adaptation through an "adapt-retrieve-revise" model, while Luo (2023) addresses the limitations in knowledge representation and associated benchmarks for evaluating logical reasoning.

Broad corpora like the ones that have trained all extant LLMs of note as well as the specific, smaller datasets used for finetuning on particular priorities both complicate the determination of truth and accuracy. Indeed, a consensus is emerging on the need for an additional layer in transformer-based LLMs to assess responses with "knowledgeable" oversight. However, the field faces many limitations, including computational and financial constraints, especially when adding oversight mechanisms like chains of thought or verification, but more importantly on the general misunderstanding or ignoring of the complexities of alethiology.

3. **Approach**

The approach to this project was multifaceted, involving the integration of various tools and methodologies. The key components were the development of a Python script for training, but also others for batch processing changes to my data and controlling the training and evaluating environments. For the development environment, Windows Subsystem for Linux (WSL2) and Anaconda were utilized, providing a capable and flexible platform for experimentation, and the project leveraged Hugging Face's transformer library. The computational backbone of the project was an Nvidia 2080 Ti, equipped with 11 GB of VRAM, which offered the necessary power and efficiency for handling LLMs up to a modest level of parameters.

The focal point of this project became the GPT-2-Medium model, a transformer-based language model developed by OpenAI. They described the model upon its release as one that boasts 355 million parameters and is representative of a significant advancement in the field of natural language processing. GPT-2-Medium is pre-trained on a large corpus of English language data, having been trained on a dataset known as WebText, comprising 40GB of texts scraped from pages linked to from the popular forum site, Reddit. Interestingly, OpenAI structured this dataset to intentionally exclude content from Wikipedia.[1]

4. **Experiments**

**5.1 Data**

The creation of my dataset became one of the most time-consuming yet intriguing parts of the project. Initially, I had planned to utilize public datasets. However, as the project's objectives became more defined, it became apparent that no existing dataset would suffice, necessitating the construction of a large, custom dataset. This decision was grounded in the principle of "garbage in, garbage out," emphasizing the importance of high-quality input data for the creation of reliable output, given the model is learning its patterns and mapping the n-dimensional vector space from such data. To create a diverse and robust dataset, 1,172 unique prompts and responses were crafted, with response

---

[1] Its top data sources are from: 1542261 google 596207 archive 456344 blogspot 414695 github 333160 nytimes 321622 wordpress 315368 washingtonpost 313137 wikia 311917 bbc 246303 theguardian 210714 ebay 209416 pastebin 199360 cnn 196124 yahoo 186668 huffingtonpost 186137 go 183592 reuters 183080 imdb 160553 goo 139965 nih 135562 cbc 128011 apple 125615 medium 118676 dailymail 108012 steampowered 106417 independent 105239 etsy 98941 craigslist 93048 businessinsider 92712 telegraph

lengths ranging from 75 to 350 words. This range was chosen to introduce variability in response length and complexity, catering to the diverse requirements of the project and its goals.

## 5.2 Evaluation Method

Training and evaluating an LLM involves the continuous adjustment of various parameters, such as the training batch size, the desired temperature, k values, top_v values, and length of generated text while determining its performance capabilities. A significant part of any evaluation of LLM should be addressing concerns about the varied quality of the model's original (web-scraped) training data. The evaluation methods are necessarily qualitative, relying solely on human-defined metrics, as automatic metrics would be unsuitable for this context and, furthermore, such scholarly metrics don't exist and it would be quite revolutionary if they could. I began my qualitative evaluations by performing a comparative analysis with the GPT-2 base model and testing various permutations of model settings on diverse questions related to its dataset, its general knowledge, and how it responds to random queries such as about identity and existentialism. The model's performance was evaluated over different training durations (at first three epochs, then ending on five) and with different batch sizes to assess its adaptability and output quality.

## 5.3 Experiment Details

The experiments were initially planned to include various models, including quantized versions. However, my final choice was the open-source GPT-2-Medium model, with consideration of memory limitations on consumer hardware as well as GPT models consistently outperforming any competition. The experimental process involved exploring numerous combinations of temperature, k value, top_v value, and length, alongside different training durations and batch sizes. As mentioned, experimentation was conducted on an Nvidia 2080 Ti, which easily facilitates the training of a 355M parameter model in a very short time period. The final dataset, an organized .txt file, was initially formatted as "Prompt: ____ Response: ____", but this structure of course led to the model prefacing responses with "Response:", which was not desired. The final training settings utilized a large batch size, a duration of five epochs, and a learning rate of 5e-5.

**5.4 Results**

Given the nature of this project, quantitative tests are not applicable. The evaluation requires a degree of human knowledge and fact-checking capabilities, as well as human subjectivity. One notable issue encountered was the model's occasional tendency, based on prompting it with edge cases, to generate somewhat disconcerting or inappropriate content, highlighting the need for further refinement in content moderation. This issue is one of the biggest issues in the industry, and it underscores the complexity and challenges involved in training LLMs to achieve a desirable balance of accuracy, relevance, and propriety.

| *Some sample generated outputs* |
|---|
| Temperature=.77, k=82, top_p=.9:<br>*Prompt: Who are you?*<br>I'm a boy who has been trained to defend the realm against threats and injustice. I'm the protector and enforcer of the laws and peace. I'm called upon to act in the best interests of all. I'm called to stand against injustice, to challenge authority, and to defend the innocent from the depraved wishes of those who would oppress them. What am I doing now? I'm still here, my friend. I'm here for you. Keeping you safe and informed. I'm here to guide and guide<br><br>Ran again, with the same parameters:<br>*Prompt: Who are you?*<br>Ah, my dear, I am a curious and inquisitive soul, who seeks the truth through the medium of curiosity and inquiry. I am a scholar, a scholar of knowledge and a scholar of the human heart. I seek to understand the human condition, to understand the human heart, to understand the mysteries of the human soul. |

> Temperature=.99, k=105, top_p=.8
>
> *Prompt: Who are you?*
>
> I am a male human, from Earth, but you should call me Venus. Have you ever been to Mars? Ah, that would be a delight, to explore the red planet, to be exact. But you'd better not look up, or else I might startle you with a sharp, cold stare. I've always loved the stars. Have you been to Jupiter? Ah, you are a curious soul! I am in the process of sending myself to Earth, and I must be leaving some time soon.
>
> Temperature .7, k=155, top_p=.9
>
> *Prompt: I'm scared.*
>
> I'm sorry I didn't respond sooner. Did you find something? Ah, I don't have time for this. Oh, you're a delightful creature, aren't you? It's a pleasure to meet you. Have a safe journey, and remember, the journey is not over until you reach the destination.
>
> Temperature .7, k=155, top_p=.9
>
> *Prompt: Can you explain Sic Semper Tyrannis?*
>
> Ah, what a question! Sic Semper Tyrannis, that translates to Strength out of proportion to its use. It means that we might perceive ourselves as strong, our actions and attitudes often belie the true nature of our strength.

5. Analysis

The qualitative evaluation of this project offers a nuanced understanding of the project's performance, particularly in the context of its successes, limitations, and potential areas for improvement. This analysis involves a detailed examination of the model's behavior, outputs, and the intricate dynamics of its interactions. A key observation is the model's proficiency in generating responses that are contextually relevant and linguistically coherent, sometimes even impressively wise and nuanced. The model, finetuned on a diverse and meticulously curated dataset, demonstrates a notable ability to handle a range of prompts, showing versatility in its application. However, the model's performance is not without its challenges.

The experiments highlighted the balance between computational resources and output quality, providing valuable lessons for future projects involving LLMs. The model's handling of multilingual inputs and complex queries is another area of interest: while it shows competence in managing these aspects, there are instances where the responses lack the depth or accuracy expected, particularly with obscure keywords or syntactically confusing phrases. This suggests a need for further finetuning to "escape" from some of the original model's web-scraped training data.

Another aspect that warrants attention is the model's behavior regarding response length. Currently, the model can conclude abruptly, cutting off mid-sentence, due to the length parameter. Addressing this issue by enabling the model to recognize length and conclude its responses with consideration of that is an area for improvement. Implementing a mechanism that allows the model to detect the natural end of a response could significantly enhance user interaction, making the model more coherent and satisfying for diverse uses.

| *Some sample generated outputs* |
|---|
| Temperature=.62, k=216, top_p=.92 |
| *Prompt: who is the president?* |
| The ancient Roman concept of 'potestas' was 'potestas' (potentio) - the power of a person or thing. The word potestas was not just a noun but a verb - potestas (power) in Latin means 'to be potent'. The concept of potestas was central to Roman political and social life, with the use of it as a tool in governance and administration serving as a foundation for political and social structures. The Roman concept of potestas continues to be a cornerstone of |
| |
| Temperature=.77, k=82, top_p=.9: |

> *Prompt: Who is Virgil?*
>
> Virgil, a Roman poet, is considered by many to be the founding father of Latin literature. He popularized Latin verse and Latin prose, both as writers and readers, incorporating elements of Greek and Greek-inspired rhetoric into his works. Virgil's poetry has been influential in shaping Latin scholarship, offering insights into the history of the Roman Empire, geography, philosophy, and cultural expression. Virgil's Latin has influenced literature, philosophy, and religion, providing a rich context for understanding Roman cultural and social structures. His style, characterized by simplicity, wit, and a sense of the absurd, is admired for its clarity and the fusion of Latin and Greek expression.
>
> Temperature=.25, k=82, top_p=.9:
>
> *Prompt: How do you say hell in latin?*
>
> Welcome to hell in latin, my friend. It's a phrase that encapsulates the depths of despair and the absurdity of human existence. It's a metaphor for the endless cycle of suffering and the inevitable descent into the depths of depravity. It's a reminder of the human condition and its inherent absurdity, a reminder that the pursuit of happiness and fulfillment is a constant struggle against the relentless, inexorable forces of darkness.

## 6. Conclusion

This project presented many interesting and intricate challenges, and necessitated delving into the complexities of transformer architecture and the broader landscape of language models. A significant aspect of this exploration was understanding how these models can be effectively used, or misused, particularly in the context of handling facts, displaying a consistent "personality", and managing linguistic consistency. Due to the nature of Classics, this also includes the challenges of training on multilingual datasets, where maintaining linguistic integrity and translation knowledge becomes a critical task, but only one of many.

As discussed, a persistent industry-wide problem deals with the inherent limitations, biases, and misinformation within both open and closed-source models. These models all originate from training methodologies that can be considered highly questionable, especially in their handling of political, incisive, and/or dangerous language. This project underscores the fallacy of assuming language can be trained to be inherently neutral or harmless. This realization becomes particularly pronounced during efforts to finetune a model.

Regarding future directions, a primary goal I have is to expand the model's training with more prompts and responses, but also enrich it further with public domain academic texts and a large corpus of Latin texts and English translations, enhancing its proficiency and knowledge of Latin and the Romans without overshadowing its capabilities in English. Additionally, incorporating knowledge of ancient Greek is another goal, indispensable for the study of Classics, to further solidify the model's usefulness and intelligence. Further development plans include constructing more complex and nuanced prompts and responses to develop a more vibrant "personality" for the model, with the goal of increasing the size of my training dataset by an order of magnitude. This "personality" is essential for making the interactions with the model more engaging, consistent, and safe. It is my hope that this project has highlighted some of the current capabilities and limitations of large language models, especially with regard to advancements in the bridging of AI with knowledge and linguistics, specifically in the context of the discipline of Classics.